\documentclass[letterpaper, 10 pt, journal, twoside]{IEEEtran}
\usepackage{times}
\usepackage[pdftex]{graphicx}
\usepackage{subfigure}
\usepackage{amsmath,amssymb,amsopn,amstext,amsfonts}
\usepackage{cancel}
\usepackage{cite}
\usepackage{pdfsync}
\usepackage{balance}
\usepackage{color}
\usepackage{mathtools}
\usepackage[ruled,vlined,linesnumbered]{algorithm2e}
\usepackage{bm}

\usepackage{diagbox}
\usepackage{float}
\usepackage{epstopdf}
\usepackage{pifont}
\usepackage{fixltx2e}
\usepackage{multirow}
\usepackage{url}
\usepackage{svg}
\usepackage[linkcolor=black,citecolor=black,urlcolor=black,colorlinks=true,bookmarks=true]{hyperref}

\usepackage{algpseudocode}
\usepackage{booktabs}
\usepackage{caption}
\usepackage{multicol}
\usepackage{verbatim}

\newcommand{\tp}{^{\mathrm{T}}}

\newcommand{\df}[1]{\mathrm{d}{#1}}

\newcommand{\norm}[1]{\Vert{#1}\Vert}

\graphicspath{{./figures/}}
\DeclareGraphicsExtensions{.png,.jpg,.eps,.pdf}
\IEEEoverridecommandlockouts

\begin{document}
	
	\title{\LARGE \bf Enhanced Decentralized Autonomous Aerial Robot Teams with \\ Group Planning}
	
	\author{Jialiang Hou$^{1, 3}$, Xin Zhou$^{2, 3}$, Zhongxue Gan$^{1}$, and Fei Gao$^{2, 3}$ 
		
		\thanks{Manuscript received: February, 24, 2022; Revised: April, 26, 2022;
			Accepted: June, 22, 2022.}
		\thanks{This paper was recommended for publication by Editor Hsieh, M. Ani upon evaluation of the Associate Editor and Reviewers' comments. This work was supported by Shanghai Municipal Science and Technology Major Project 2021SHZDZX0103, National Natural Science Foundation of China under Grant 62003299, the Shanghai Engineering Research Center of AI \& Robotics, Fudan University, China, and the Engineering Research Center of AI \& Robotics, Ministry of Education, China. (Corresponding author: Fei Gao, Zhongxue Gan.)}
		
		\thanks{$^{1}$Academy for Engineering and Technology, Fudan University, Shanghai, 200433, China. }
		\thanks{$^{2}$State Key Laboratory of Industrial Control Technology, Institute of Cyber-Systems and Control, Zhejiang University, Hangzhou, 310027, China.}
		\thanks{$^{3}$Huzhou Institute of Zhejiang University, Huzhou, 313000, China.}
		\thanks{Email: \{jlhou19, ganzhongxue\}@fudan.edu.cn; fgaoaa@zju.edu.cn}
		\thanks{Digital Object Identifier (DOI): see top of this page.}}

	\maketitle

	\begin{abstract}
		
		Designing autonomous aerial robot team systems remains a grand challenge in robotics.
		Existing works in this field can be categorized as centralized and decentralized. 
		Centralized methods suffer from scale dilemmas, while decentralized ones often lead to poor planning quality. 
		In this paper, we propose an enhanced decentralized autonomous aerial robot team system with group planning. 
		According to the spatial distribution of agents, the system dynamically divides the team into several groups and isolated agents. 
		For conflicts within each group, we propose a novel coordination mechanism named group planning. The group planning consists of efficient multi-agent pathfinding (MAPF) and trajectory joint optimization, which can significantly improve planning quality and success rate. We demonstrate through simulations and real-world experiments that our method not only has applicability for a large-scale team but also has top-level planning quality.
		
	\end{abstract}
	
	
	\IEEEpeerreviewmaketitle
	
\section{INTRODUCTION}
\IEEEPARstart{D}{esigning} robot team systems remains a grand challenge in robotics and will have great advances and impacts in the next 5 to 10 years \cite{yang2018grand}.
Aerial robot teams, as a popular topic in the team community, can solve many challenges faced by human civilization, such as natural disasters, space colonization, and air traffic.

Judging by the way to deploy computation and communication resources, the planning of aerial robot teams can be categorized as centralized and decentralized methods.
Centralized methods \cite{mellinger2012mixed,augugliaro2012generation,honig2018trajectory,park2020efficient} simultaneously solve the planning problem for all agents and then allocate
the plans to each one.
However, the interaction among agents makes the complexity grow combinatorially. 
Also, it is impractical to share information across the large-scale robot teams online.
The abovementioned issues prevent the application of centralized methods to large robot teams in real world. 
Recently, our community witnesses the emergence of decentralized methods \cite{tordesillas2021mader,zhou2020ego,zhou2021decentralized}. Although they can be applied to larger-scale teams by amortizing the computation and communication, the planning only uses single agent's environmental information. Therefore, the planning quality and success rate deteriorate as scale increases, especially in obstacle-dense environments.

\begin{figure}[t]
	\centering
	\includegraphics[width=1.0\linewidth]{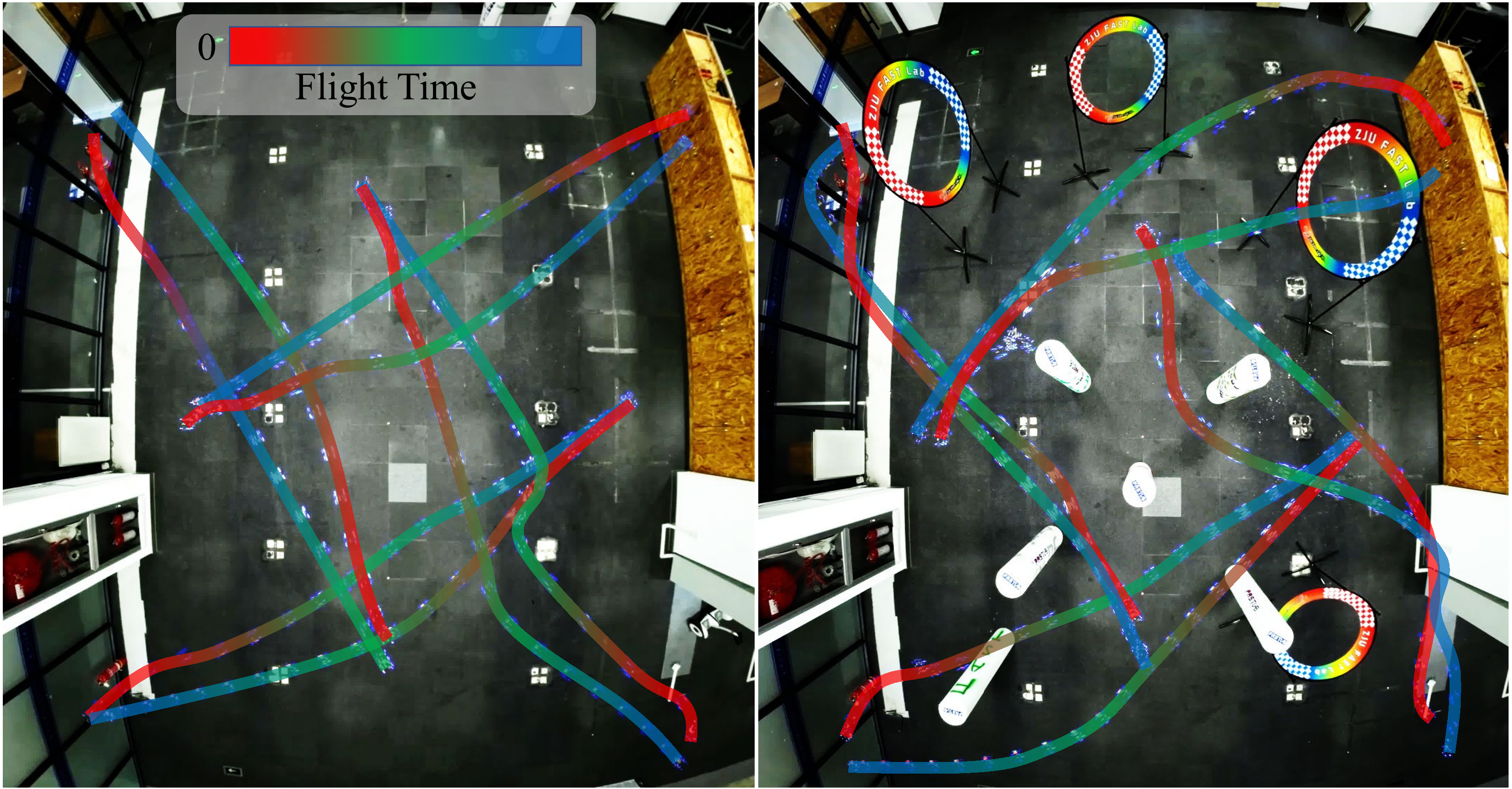}
	\captionsetup{font={small}}
	\caption{Indoor experiments.}
	\label{pic:indoor}
	\vspace{-0.4cm}
\end{figure}

\begin{figure}[t]
	\centering
	\includegraphics[width=1.0\linewidth]{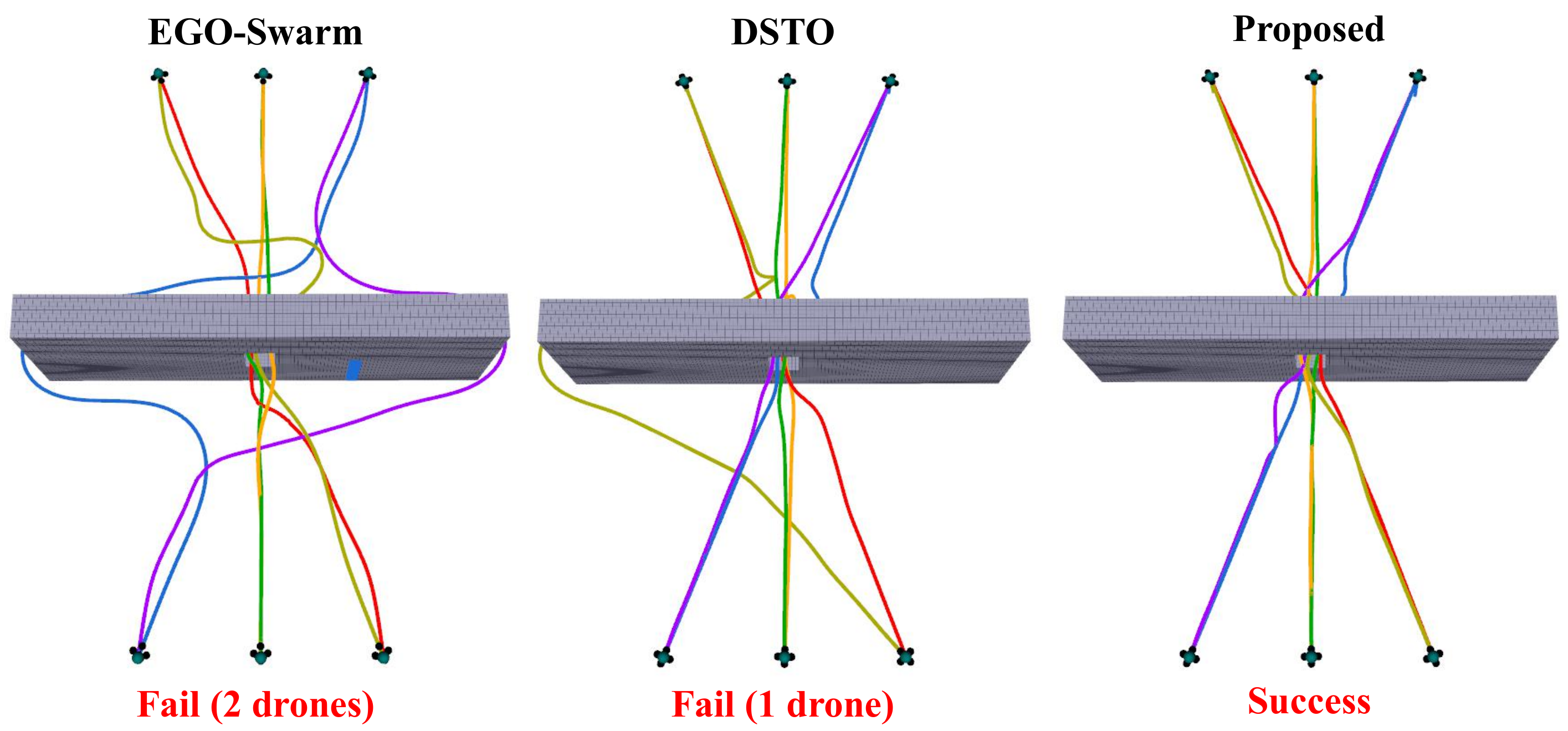}
	\captionsetup{font={small}}
	\caption{Passing through a 0.8x1.5m gate with EGO-Swarm, DSTO and the proposed method.}
	\label{pic:sim_ben}
	\vspace{-1.6cm}
\end{figure}

Investigating the above issues, we propose an enhanced decentralized aerial robot team system with group planning. 
This paper tackles the conflict between efficiency and quality, balances the coordination of individuals and groups, and introduces a robust and flexible framework (in Fig. \ref{pic:system_framework}) for aerial robot teams. 
We consider the key challenges for building an applicable large-scale aerial robot team to be twofold.
There is a need for a powerful trajectory planning backbone, as well as a flexible grouping mechanism to solve potential conflicts originating from agents and environments. 
When agents cluster in a narrow space, we dynamically form several groups based on their spatial distribution, while regarding each agent as an independent individual when they disperse.
As for groups, we propose a method called \textit{group planning}, as a portable plugin for fully decentralized methods. It conducts trajectory planning utilizing complete information of multiple agents where hard-to-resolve inter collisions may happen.
We first implement an online multi-agent pathfinding method to generate multiple collision-free paths. Based on these paths, we propose a trajectory optimization method such that all trajectories can converge to the joint optimum.
Exploiting our \textit{group planning} strategy, the proposed system is a significantly enhanced decentralized one over the solution in \cite{zhou2021decentralized}.


We compare our method with state-of-the-art methods. The results show that our method can generate the shortest flight time and distance of trajectory with the fewest replan times. 
In addition, we implement a systematic solution for decentralized autonomous aerial robot teams, including onboard perception, planning, control, and communication.
The contributions of this paper are summarized as follows:

1) A more efficient three-dimensional online multi-agent pathfinding method, combined with the ECBS\cite{Barer2014SuboptimalVO}, for finding multiple collision-free paths that are close to the optimal trajectories.

2) An enhanced decentralized autonomous aerial robot team system with group planning where joint optimization is conducted on demand to significantly improve planning quality and success rate.

3) Open-source code\footnote{https://github.com/ZJU-FAST-Lab/EDG-TEAM} of our system that is extensively validated by simulations and real-world experiments.

\section{RELATED WORK}

\subsection{Multi-Agent Pathfinding}

\begin{table}[h]
	\centering
	\caption{Evaluation of MAPF algorithms}
	\label{tab:mapf_eva}
	\renewcommand{\arraystretch}{1.35}
	\begin{tabular}{|c|c|c|c|c|}
		\hline
		\textbf{Method} &
		\textbf{Optimality} &
		\textbf{Completeness} &
		\textbf{Scalability} &
		\textbf{\begin{tabular}[c]{@{}c@{}}Run\\  Time\end{tabular}} \\ \hline
		CA*           & Sub-optimal & Incomplete & \multirow{2}{*}{Very Small} & \begin{tabular}[c]{@{}c@{}}Very\\  Slow\end{tabular}          \\ \cline{1-3} \cline{5-5} 
		\begin{tabular}[c]{@{}c@{}}A*+\\ ID/OD\end{tabular} &
		\multirow{4}{*}{Optimal} &
		\multirow{5}{*}{\textbf{Complete}} &
		&
		\multirow{4}{*}{Slow} \\ \cline{1-1} \cline{4-4}
		M*            &             &            & \multirow{3}{*}{Small}      &                                                               \\ \cline{1-1}
		ICT           &             &            &                             &                                                               \\ \cline{1-1}
		CBS           &             &            &                             &                                                               \\ \cline{1-2} \cline{4-5} 
		\textbf{ECBS} & Sub-optimal &            & \textbf{Very Large}         & \textbf{\begin{tabular}[c]{@{}c@{}}Very\\  Fast\end{tabular}} \\ \hline
	\end{tabular}
\end{table}

The MAPF problem is to find a spatial-temporal collision-free path for each agent in a group, given prescribed start and goal positions. Table \ref{tab:mapf_eva} summarizes the properties of the search-based methods corresponding to MAPF.

If the A* search is simply performed by joining the state spaces of all agents, the search space will grow exponentially with the number of the agents, resulting in computational intractability. 
Cooperative A* (CA*)\cite{silver2005cooperative} performs A* searches in priority order to solve the combinatorial explosion problem, which is suboptimal and incomplete.
A*+ID\cite{standley2010finding} breaks the problem down into several independent subproblems and solves each one separately. When there is a conflict between any two subproblems, the corresponding subproblems are combined to solve the conflict. The solutions of all subproblems are combined as the final solution until no conflict occurs, but the algorithm's scalability is restricted. 
A*+OD\cite{standley2010finding} specifies that only one agent can move at a time, reducing the search dimension but increasing the depth of the search tree. 
M*\cite{wagner2011m} dynamically changes the search dimension based on whether or not a conflict occurs. When there is no conflict, each agent takes the best action possible on its own; otherwise, the local search dimension is increased during the conflict. 
Conflict-based methods address the problem on two levels. The high level converts conflicts into constraints added to agents, while the low level performs a search. 
ICTS algorithm\cite{sharon2013increasing} searches on an increasing cost tree (ICT). 
CBS algorithm\cite{sharon2015conflict} performs searching on a constraint tree (CT). When there is a conflict, the high level creates a new node and converts the conflict into a constraint added to the agents, while the low-level agents perform A* searches. When all agents reach their respective targets with no conflict, the node is the final solution. The CBS is optimal and complete, but it is more time-consuming to solve. 
ECBS algorithm\cite{Barer2014SuboptimalVO} sacrifices a small amount of solution quality in exchange for greater efficiency and scalability.

The methods mentioned above have limitations, and there is no universal winner. The paper uses MAPF to generate the initial paths and provide an initial value for the back-end optimization. We prefer algorithms with high completeness, scalability, and efficiency, even slightly sacrificing some optimality. 
We choose the ECBS algorithm as the front-end of group planning based on comparisons. Because most current MAPF implementations are in a two-dimensional or offline three-dimensional environment\cite{Barer2014SuboptimalVO,honig2018trajectory,park2020efficient}, we implement the ECBS algorithm in an online three-dimensional gridmap and add a tie breaker function to improve search efficiency and solution quality.

\subsection{Aerial Robot Teams}

For centralized strategies, the methods proposed by Mellinger and Augugliaro\cite{mellinger2012mixed,augugliaro2012generation} can generate feasible trajectories for small aerial robot teams in a few seconds. However, they hardly handle large aerial robot teams and have limited application scope.
Honig and Park\cite{honig2018trajectory,park2020efficient} use B-spline or B\'ezier to generate safe and dynamically feasible trajectories for aerial robot teams in an offline known environment. The convex hull of the B-spline or the B\'ezier compresses the solution space, preventing a trajectory from being aggressive near its physical limits. Similar to ours, they also use the ECBS algorithm for a front-end search. However, our implementation is more efficient, and the generated paths are closer to the optimal trajectories. 

\begin{figure*}[ht]
	\centering
	\includegraphics[width=0.9\linewidth]{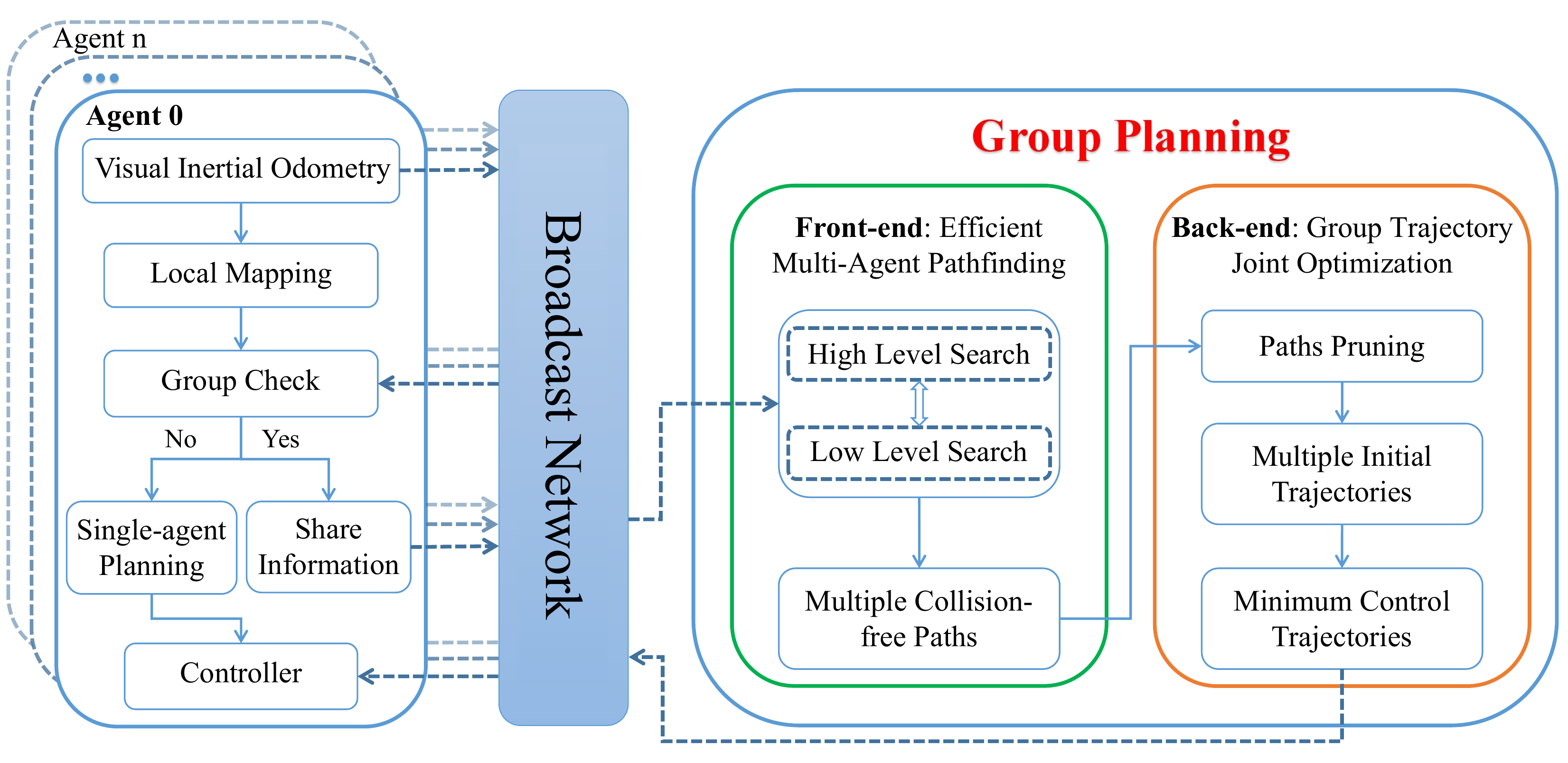}
	\captionsetup{font={small}}
	\caption{Enhanced Aerial Robot Team Framework}
	\label{pic:system_framework}
	\vspace{-0.5cm}
\end{figure*}

For decentralized strategies, 
Tordesillas et al.\cite{tordesillas2021mader} propose a decentralized asynchronous solution that employs MINVO basis to alleviate the issue of compressed solution space. However, the work is only validated through simulation.
Zhou et al.\cite{zhou2020ego} propose a decentralized and asynchronous systematic solution that can be used in real world. Due to the use of B-spline, optimizing the time term is difficult, which results in twisted trajectories when multiple drones cluster. 
To solve the problem of the trajectory twist, Zhou et al.\cite{zhou2021decentralized} propose a spatial-temporal optimization solution based on MINCO\cite{wang2021geometrically}. However, the planning quality and success rate decrease dramatically when multiple drones cluster. The proposed method solves the problem.
 
\section{Enhanced Aerial Robot Team Framework}
\textbf{Group Planning Criteria}~ \textit{A team contains $n$ agents, and the distance between the agents$\{a_i, a_j\}$ is $d_{ij}$, if and only if
	\begin{gather}
	n_{min} \leq n \leq n_{max}~(n_{min} \geq 2),\\
	d_{ij} \leq d_{safe} ~(1 \leq i \leq n-1,~i \textless j \leq n),
	\end{gather}
	the n agents satisfy the group planning criteria, where $n_{min}$ and $n_{max}$ are the upper and lower limits of the team number. $d_{safe}$ is the safety distance.}

The proposed framework is illustrated in Fig. \ref{pic:system_framework}. The \textit{Group Check} module receives agents' odometry data, and the agent who meets the group planning criteria switches to the group planning mode. Each agent shares the start and local target positions and local map with the core agent in the group. Because of the cooperation among these agents, we define these agents as a group. Otherwise, the agent that does not meet the group planning criteria switches to single-agent planning\cite{XinZhou2020EGOPlannerAE}.

The \textit{group planning} consists of front-end \textit{efficient multi-agent pathfinding} (Section \ref{sec:path}) and back-end \textit{group trajectory joint optimization} (Section \ref{sec:traj}). The \textit{efficient multi-agent pathfinding} performs two-level searches based on obstacles and conflicts among agents to generate multiple collision-free paths. Based on these paths, the \textit{group trajectory joint optimization} generates multiple optimal trajectories in a coarse-to-fine process. It first prunes the paths and creates the intial trajectories as the initial value of the optimization problem. Then, these trajectories are jointly optimized. Finally, multiple safe, dynamically feasible, and minimum control trajectories\cite{wang2021geometrically} are executed.



\section{Efficient Multi-agent Pathfinding}
\label{sec:path}


This section aims to find a collision-free path closer to the optimal trajectory for each agent in the group. For the first time, we implement a more efficient online multi-agent pathfinding (EMAPF) method on a three-dimensional (3D) gridmap based on ECBS\cite{Barer2014SuboptimalVO}.


Like ECBS, our method performs high-level and low-level searches in Algorithm 1. The two level searches use focal search ($f_1, f_2$)\cite{Barer2014SuboptimalVO}, where $f_1$ and $f_2$ have different meanings at two levels. The focal search contains two lists OPEN and FOCAL, which can limit the maximum cost of the solution to $\omega*C^*$ ($\omega$ is the suboptimal factor. $C^*$ is the minimum cost of the node in OPEN.). $f_1$ determines the nodes in FOCAL. FOCAL is subset of OPEN that includes all nodes n in OPEN where $f_1(n) \leq \omega * f_{1_{min}}$ ($f_{1_{min}}$ is the minimal value of $f_1$.). $f_2$ determines which node in FOCAL should be expanded.


In high level, the search is performed on a Conflict Tree (CT) consisting of multiple nodes. The solution of a node contains the paths of all the agents$\{a_1, ..., a_i, ..., a_k\}$ in the group. If a conflict exists among the paths in the extended node, it is converted into a low-level constraint to form a new node. The search terminates until we get a collision-free node containing all agents' collision-free paths in the group. In low level, the agent performs constraint-based pathfinding, and the node represents the agent's position. The specific implementation of the two level searches in our method is detailed as follows.

\subsection{High-level Search}

High-level search applies focal search $(f_1, f_2)$ on a CT. The goal is to find a conflict-free node. In high level, $f_1$ is the cost of a CT's node, and $f_2$ is an inadmissible heuristic function that represents the number of the CT's node conflicts. The lower bound ($LB$) and the FOCAL of the CT are as follows:
\begin{equation}
LB = min (LB(n)|n \in OPEN),
\end{equation}
\begin{equation}
FOCAL = \{n|n\in OPEN, n.cost \leq LB \cdot \omega\},
\end{equation}
where $n$ is a CT's node. $LB(n)$ and $n.cost$ come from the return value of low-level search. 

\subsection{Low-level Search}

\setlength{\textfloatsep}{-15pt}
\begin{algorithm}[t]
	\label{alg:emapf}
	
	\KwIn{startPositions, goalPositions, 3D Gridmap\\
		S.constraints = $\emptyset$ // S is start node.\\
		S.solution = find paths using \textbf{lowLevelSearch()}\\
		S.cost = $\sum_{i} S.solution[i].cost$\\
		S.LB = $\sum_{i} S.solution[i].f_{min}$\\}
	\KwOut{Goal node solution}
	
	Insert S to OPEN\\
	Insert S to FOCAL\\
	\While{OPEN not empty}{
		\If{the lowest solution cost increase in OPEN}{expand FOCAL from OPEN}
		N $\leftarrow$ lowest conflict node from FOCAL\\
		\If{N not conflict}{\Return{N.solution} // N is goal node.}
		Constraints $\leftarrow$ getFirstConflict(N)\\
		\ForEach{$a_{i}$ in Constraints}{
			P $\leftarrow$ new node\\
			P.constraints $\leftarrow$ N.constraints + ($a_i$,s,t)\\
			P.solution = N.solution\\
			Update P.solution by invoking \textbf{lowLevelSearch($a_i$)}\\
			P.cost = $\sum_{i} P.solution[i].cost$\\
			P.LB = $\sum_{i} P.solution[i].f_{min}$\\
			Insert P into OPEN\\
			\If{P.cost satisfies the condition of FOCAL}{Insert P in FOCAL}
		}
	}	
	\caption{Efficient Multi-Agent Pathfinding}
	
\end{algorithm}

\begin{figure}[h]
	\centering
	\includegraphics[width=1.0\linewidth]{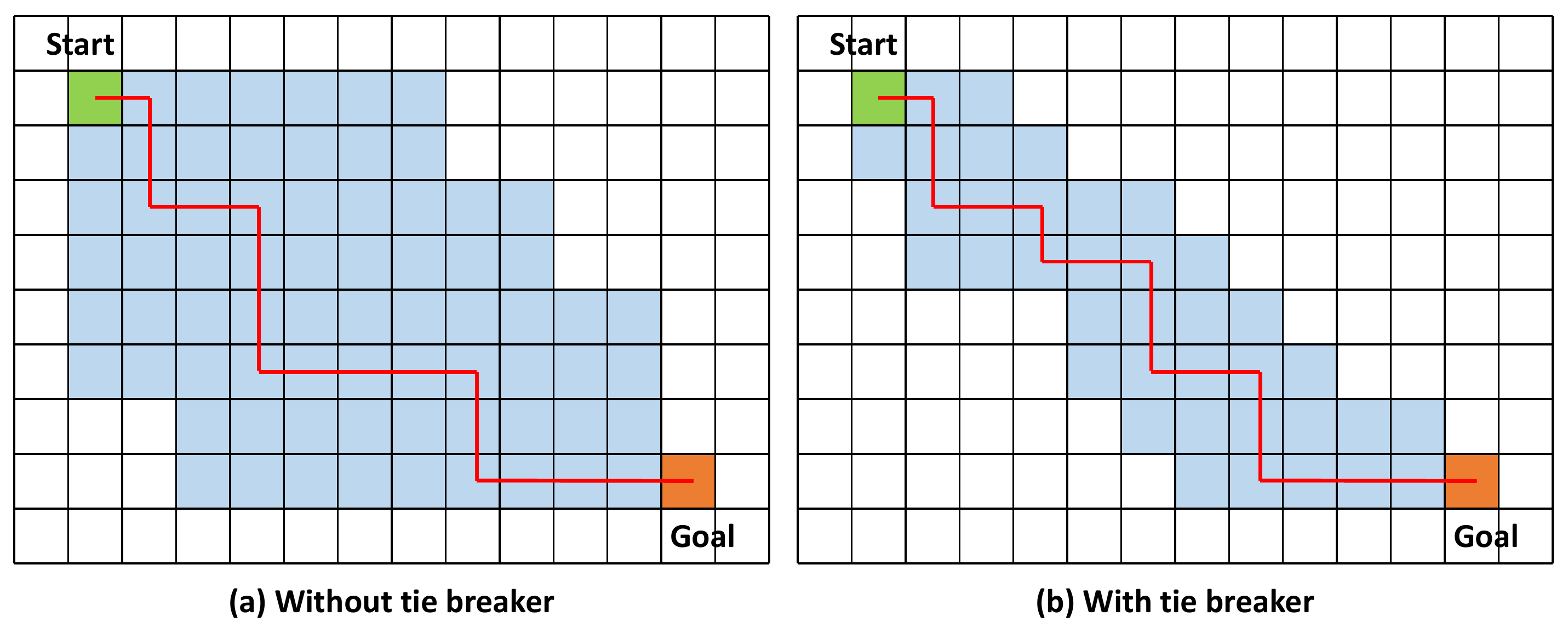}
	\captionsetup{font={small}}
	\caption{The impact of tie breaker on traversal space; The light blue area indicates traversal space. The red line indicates the path.}
	\label{pic:traversal_space_comparison}
	\vspace{-0.1cm}
\end{figure}

Low-level search applies focal search $(f_1, f_2)$ to find a path for each agent $a_i$. In low level, we set $f_1 = g + h + t$, where $g$ is the current best estimate of the accumulated cost from the start node to the current node, $h$ is the estimated lowest cost from the current node to the goal node. Function $t$ is the distance from the current traversed node to the connection line between the start and goal nodes, called the $tie~breaker$ function. $f_2$ is an inadmissible heuristic function representing the number of conflicts in the current path. 
The purposes of designing the $tie~breaker$ function $t$ are: (1) to reduce the traversal space; In Fig. \ref{pic:traversal_space_comparison}, if $f_1$ is set to $f = g + h$ as in the $A^*_\epsilon$ algorithm \cite{Pearl1982StudiesIS}, there will be multiple nodes with the same $f$ value when searching the path, which would result in a larger traversal space; (2) to make the searched path closer to the optimal trajectory.
The entire search is performed on a real-time updated 3D gridmap, and it only takes $O(1)$ time complexity to determine whether the traversed node is occupied.


A lower bound on the cost of the path for agent $a_i$ is $f_{min}(i)$ in the low-level search. Therefore, we can construct the lower boundary $LB(n)$ and the cost of the current CT's node $n$ in high-level search as follows:
\begin{gather}
LB(n) = \sum_{i=1}^k f_{min}(i),\\
n.cost = \sum_{i=1}^k cost(i),
\end{gather}
where $k$ is the number of agents in the group. The low-level search will return to the high-level search with newly updated $LB(n)$ and $n.cost$ of the CT's node $n$. 


In summary, our method reduces traversal space during the search process while efficiently checking the information about obstacles. The obtained paths are close to the optimal trajectories (see Fig. \ref{pic:ecbs}), providing a good initial value for trajectory optimization. The quantitative comparison is shown in the TABLE \ref{tab:comparisons}.

\section{Group Trajectory Joint Optimization}
\label{sec:traj}


This section aims to make all agents' trajectories in the group converge on a joint optimum. Therefore, we jointly optimize all trajectories. However, it is challenging to achieve reciprocal avoidance among dynamically changing trajectories. We address the problem in Section \ref*{sec:swarm_avoid}. The trajectory representation and the joint trajectory optimization problems are detailed as follows.

\subsection{MINCO Trajectory Class} 

In this paper, we adopt $\mathfrak{T}_{\mathrm{MINCO}}$\cite{wang2021geometrically} for trajectory representation, which is a minimum control effort polynomial trajectory class defined as
\begin{align*}
\mathfrak{T}_{\mathrm{MINCO}} = \Big\{&p(t):[0, T]\mapsto\mathbb{R}^m \Big|~\mathbf{c}=\mathcal{M}(\mathbf{q},\mathbf{T}),~\\
&~~\mathbf{q}\in\mathbb{R}^{m(M-1)},~\mathbf{T}\in\mathbb{R}_{>0}^M\Big\},
\end{align*}
where $\mathbf{q}=(\mathbf{q}_1,\dots,\mathbf{q}_{M-1})$ denotes the intermediate waypoints and $\mathbf{T}=(T_1, T_2, \dots, T_M)\tp$ denotes the duration for all pieces. p(t) is an $m$-dimensional $M$-piece polynomial trajectory with degree $N = 2s - 1$, $s$ is the order of integrator chain. The $i$-th piece of $p(t)$ is defined by\\
\begin{equation}
p_i(t)=\mathbf{c}_i\tp\beta(t),~\forall t\in[0,T_i],
\end{equation} 
where $\mathbf{c} = (\mathbf{c}_1\tp, \dots, \mathbf{c}_M\tp)\tp \in \mathbb{R}^{2Ms \times m}$ the polynomial coefficient, $\mathbf{c}_i\in\mathbb{R}^{2Ms \times m}$ is the cofficient matrix of the $i$-th piece.$\beta(t)=[1, t, \cdots, t^N]\tp$ is the natural basis.

All trajectories in $\mathfrak{T}_{\mathrm{MINCO}}$ are compactly parameterized by only $q$ and $T$. Evaluating an entire trajectory from $q$ and
$T$ can be done via the linear-complexity
formulation:
\begin{equation}
\label{eq:mapping}
\mathbf{c}=\mathcal{M}(\mathbf{q},\mathbf{T}).
\end{equation}

The task-specific second-order continuous penalty functions $F(\mathbf{c},\mathbf{T})$ with available gradients are applicable to $\mathfrak{T}_{\mathrm{MINCO}}$ trajectories. The corresponding objective of $\mathfrak{T}_{\mathrm{MINCO}}$ is computed as:
\begin{equation}
J(\mathbf{q},\mathbf{T})=F(\mathcal{M}(\mathbf{q},\mathbf{T}), \mathbf{T}).
\end{equation}
The mapping Eq.\ref{eq:mapping} gives a linear-complexity way to compute $\partial J/\partial\mathbf{q}$ and $\partial J/\partial\mathbf{T}$ from the corresponding $\partial F/\partial\mathbf{c}$ and $\partial F/\partial\mathbf{T}$. Sequentially, a high-level optimizer can optimize the objective efficiently.

\subsection{Joint Optimization Problem Formulation} 

The basic requirements for trajectories in the group include safety, smoothness, and dynamical feasibility. Meanwhile, it is preferable to minimize control effort cost and execution time of all trajectories. 

We adopt the compact parameterization of $\mathfrak{T}_{\mathrm{MINCO}}$, temporal constraint elimination, and constraint penalty to transform trajectories generation problem into an unconstrained nonlinear optimization problem:
\begin{equation}
\label{eq:objfun}
\min_{\cup_{K}\mathbf{q},\mathbf{T}}~ \sum_{K}{\lambda\cdot{[J_e, J_t, J_d, J_o, J_w, J_u]}},
\end{equation}
where K is the number of agents in the group, $\lambda$ is the weight vector.

\subsubsection{Control Effort $J_e$}
The control effort cost of the k-th agent trajectory and the gradients for its $i$-th piece are:
\begin{gather}
J_e = \sum_{i}{\int_{0}^{T_i}{\norm{p^{(s)}_i(t)}^2\df{t}}},\\
\frac{\partial J_e}{\partial \mathbf{c}_i}=2 \left( \int_{0}^{T_i} \beta^{(s)}(t)\beta^{(s)}(t)^{\tp} dt \right) \mathbf{c}_i,\\
\frac{\partial J_e}{\partial T_i}=\mathbf{c}_i\tp\beta^{(s)}(T_i)\beta^{(s)}(T_i)^{\tp}\mathbf{c}_i.
\end{gather}

\subsubsection{Temporal Constraint Elimination}
An open-domain constraint is $\mathbf{T} \succ \mathbf{0}$, which is directly eliminated by variable transformations as is done in\cite{wang2021geometrically}:
\begin{equation}
\label{eq:TDiffTransformation1}
T_i=e^{\tau_i},
\end{equation}
where $\tau_i$ is the unconstrained virtual time. 

\textit{Execution Time Cost $J_t$:}
The execution time cost $J_t=\sum_{i=1}^{M} T_i$ and its gradient ${\partial J_t}/{\partial \mathbf{c_i}} = \mathbf{0}$, ${\partial J_t}/{\partial \tau_i} = ({\partial J_t}/{\partial T_i}) e^{\tau_i}$.

\subsubsection{Penalty for Continuous-Time Constraints} 
Continous-time constraints $\mathcal{G}(p(t),\dots,p^{(s)}(t)) \preceq \mathbf{0}, \forall t\in[0,T]$ contain infinite inequality constraints that cannot be directly solved using constrained optimization. We transform $\mathcal{G}$ into finite inequality constraints using integral of constraint violations\cite{Jennings1990ACA}. The penalty with gradient for k-th agent trajectory can be derived:
\begin{subequations}
\label{eq:PieceTimeIntegralPenalty}
\begin{gather}
\allowdisplaybreaks[4]
J_\Sigma(\mathbf{c},\mathbf{T})=\sum_{i=1}^{M}{J_i(\mathbf{c}_i,T_i,\kappa_i)},\\
J_i(\mathbf{c}_i,T_i,\kappa_i)=\frac{T_i}{\kappa_i}\sum_{j=0}^{\kappa_i}\bar{\omega}_j\chi^\mathrm{T}\max{(\mathcal{G}(\mathbf{c}_i,T_i,\frac{j}{\kappa_i}),\mathbf{0})^3},\\
\frac{\partial J_\Sigma}{\partial \mathbf{c}_i} = \frac{\partial J_\Sigma}{\partial \mathcal{G}} \frac{\partial \mathcal{G}}{\partial \mathbf{c}_i},~
\frac{\partial J_\Sigma}{\partial T_i} = \frac{J_i}{T_i} + \frac{\partial J_\Sigma}{\partial \mathcal{G}} \frac{\partial \mathcal{G}}{\partial t} \frac{\partial t}{\partial T_i},\\
\frac{\partial J_\Sigma}{\partial \mathcal{G}} = 3 \frac{T_i}{\kappa_i}\sum_{j=0}^{\kappa_i}\bar{\omega}_j\max{(\mathcal{G}(\mathbf{c}_i,T_i,\frac{j}{\kappa_i}),\mathbf{0})^2\circ\chi},
\end{gather}
\end{subequations}
where $\kappa_i$ is the sample number on the i-th piece, $\bar{\omega}_j$ the quadrature coefficients from the
trapezoidal rule\cite{WilliamHPress1996NumericalRI}, $\chi\in\mathbb{R}^{n_g}_{\geq0}$ is a vector of penalty weights. We define the points determined by $\{\mathbf{c}_i,T_i,{j}/{\kappa_i}\}$ as \textit{constraint points} $\mathring{\mathbf{p}}_{i,j} = p_i(({j}/{\kappa_i})T_i)$ with the $i$-th piece $p_i(t)$.


\paragraph{Dynamical Feasibility Penalty $J_d$}
According to the limit of agent maximum velocity $v_m$, acceleration $a_m$, jerk $j_m$, constraints of dynamic are denoted as
\begin{equation}
\label{eq:Jd_cost}
\mathcal{G}_v = \dot{p}(t)^2 - v_m^2,~\mathcal{G}_a = \ddot{p}(t)^2 - a_m^2,~\mathcal{G}_j = \dddot{p}(t)^2 - j_m^2.
\end{equation}
The corresponding gradients are
\begin{equation}
\label{eq:Jd_gradient}
\frac{\partial \mathcal{G}_x}{\partial \mathbf{c}_i}=2{\beta^{(n)}(t)} {p^{(n)}}(t)\tp, ~~\frac{\partial \mathcal{G}_x}{\partial t}=2{\beta^{(n+1)}}(t)\tp \mathbf{c}_i p^{(n)}(t),
\end{equation}
where $x=\{v,a,j\}$, $n=\{1,2,3\}$, and $t=jT_i/\kappa_i$.
By substituting Eq. \ref{eq:Jd_cost} \ref{eq:Jd_gradient} into Eq. \ref{eq:PieceTimeIntegralPenalty} we get the penalty $J_d$ and the gradient about $\mathbf{c}_i$ and $\mathbf{T}_i$.

\paragraph{Obstacle Avoidance Penalty $J_o$}
We adopt collision evaluation from Zhou et al\cite{XinZhou2020EGOPlannerAE}, which defines the distance from agent to obstacle as $d(p(t))$. To enforce the safety requirement, we formulate a collision penalty, which is triggered when the distance to obstacles is less than a safe clearance $\mathcal{C}_o$. The obstacle avoidance constraint and its gradient are as follows
\begin{gather}
\mathcal{G}_{o}(p(t)) = \mathcal{C}_o - d(p(t)),\\
\frac{\partial \mathcal{G}_{o_k}}{\partial c_{i}} =  -\beta(t) \mathbf{v}\tp , ~~ \frac{\partial \mathcal{G}_{o_k}}{\partial t} =  -\mathbf{v}\tp\dot{p}(t),
\end{gather}
where $\mathcal{C}_o \geq d(p(t)),0)$ and $t=jT_i/\kappa_i$.

\paragraph{Group Reciprocal Avoidance Penalty $J_w$}
\label{sec:swarm_avoid}

We jointly optimize all trajectories, which means that the trajectories of other agents change dynamically during the iterative optimization of one agent's trajectory. 
The optimization problem will not converge if we consider only the gradients about $c$ and $T$ for the current agent's trajectory. 
Therefore, in each iteration, we consider the influence of the current agent on all trajectory gradients.

In Fig. \ref{pic:swarm_avoidance}, we demonstrate the reciprocal avoidance process of the $u$-th agent's trajectory $p_u(t)$ and one of the other trajectories in the group ($k$-th agent's trajectory $p_k(t)$) at the time stamp $\tau$. $\tau$ is denoted as:

\begin{figure}[h]
	\centering
	\includegraphics[width=1.0\linewidth]{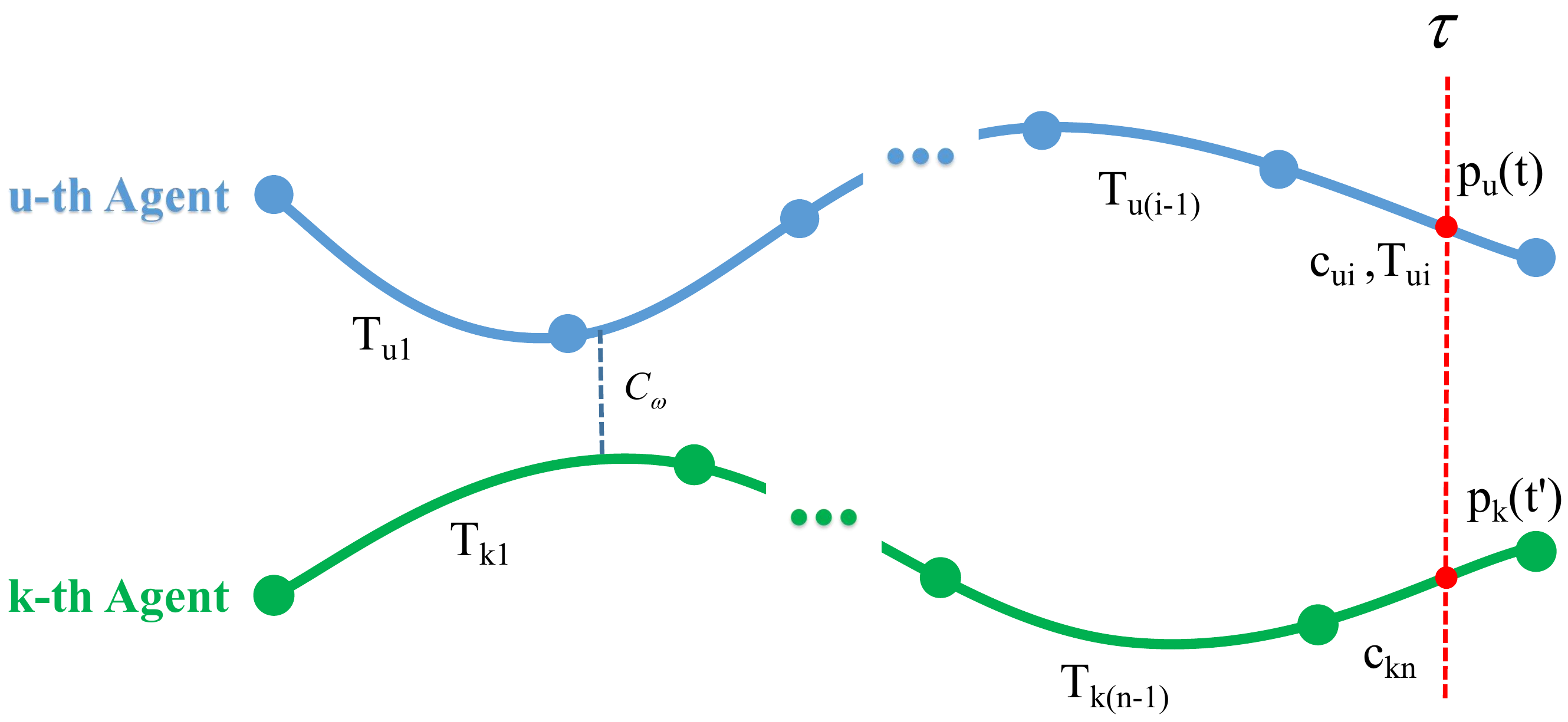}
	\captionsetup{font={small}}
	\caption{Group Reciprocal Avoidance. $c$ and $T$ indicate the coefficient and time duration of a trajectory segment. $\mathcal{C}_w$ is the safe clearance.}
	\label{pic:swarm_avoidance}
	\vspace{0.6cm}
\end{figure}

\begin{equation}
 \tau=T_{u1} + \cdots + jT_{ui}/\kappa_i = T_{k1} + \cdots + T_{k(n-1)} + t',
\end{equation}
where $t=jT_i/\kappa_i$ is the relative time of $u$-th agent. $t'$ is the relative time of $k$-th agent. $\tau$ locates in the $i$-th piece of $u$-th agent and the n-th piece of $k$-th agent. We define the constraint at the $j$-th constraint point on the $i$-th piece of $p_u(t)$ as
\begin{gather}
\mathcal{G}_w^{i,j}(p_u(t),\tau) = \left(\cdots, \mathcal{G}_{w_k}(p_u(t),\tau), \cdots\right)\tp\in\mathbb{R}^{K},\\
\mathcal{G}_{w_k}^{i,j}(p_u(t),\tau) = \begin{cases} \mathcal{C}_w^2 - d^2(p_u(t), p_k(t')) & k\neq u,\\ 0 & k=u, \end{cases} \\
d(p_u(t), p_k(t')) = \left\| \mathbf{E}^{1/2}(p_u(t) - p_k(t')) \right\|,
\end{gather}
where the matrix $\mathbf{E}$ is the downwash model of agent.

When $\mathcal{C}_w^2 \geq d^2(p_u(t), p_k(t'))$, the gradient to $\mathbf{c}_{ui}, \mathbf{c}_{kn}$ is
\begin{gather}
\label{equ:G_wk_to_cu}
\frac{\partial \mathcal{G}_{w_k}^{i,j}}{\partial \mathbf{c}_{ui}} = \begin{cases} -2\beta_{ui}(t)(p_u(t) - p_k(t'))\tp\mathbf{E} & k\neq u,\\ \mathbf{0} & k=u, \end{cases}\\
\label{equ:G_wk_to_ck}
\frac{\partial \mathcal{G}_{w_k}^{i,j}}{\partial \mathbf{c}_{kn}} = \begin{cases} 2\beta_{kn}(t')(p_u(t) - p_k(t'))\tp\mathbf{E} & k\neq u,\\ \mathbf{0} & k=u. \end{cases}
\end{gather}
The gradient to $T_{ul}$ for any $1\leq l\leq i$ can be computed as
\begin{align}
\frac{\partial J_{w}}{\partial T_{ul}}&=\sum_{k=1}^{K}\frac{\partial J_{w_k}}{\partial T_{ul}}=\sum_{k=1}^{K}\sum_{i=1}^{M}\sum_{j=0}^{\kappa_i}\frac{\partial J_{w_k}^{i,j}}{\partial T_{ul}},\\
\frac{\partial J_{w_k}^{i,j}}{\partial T_{ul}} &= \frac{J_{w_k}^{i,j}}{T_{ui}}+\frac{\partial J_{w_k}^{i,j}}{\partial \mathcal{G}_{w_k}^{i,j}} \frac{\partial \mathcal{G}_{w_k}^{i,j}}{\partial T_{ul}}, \\
\frac{\partial \mathcal{G}_{w_k}^{i,j}}{\partial T_{ul}}&=\frac{\partial \mathcal{G}_{w_k}^{i,j}}{\partial t}\frac{\partial t}{\partial T_{ul}}+\frac{\partial \mathcal{G}_{w_k}^{i,j}}{\partial t'}\frac{\partial t'}{\partial T_{ul}},\\
\frac{\partial \mathcal{G}_{w_k}^{i,j}}{\partial t} &= \begin{cases} 2\left(p_k(t') - p_u(t)\right)\tp\mathbf{E}\dot{p}_u(t) & k\neq u,\\ 0 & k=u, \end{cases} \\
\frac{\partial \mathcal{G}_{w_k}^{i,j}}{\partial t'} &=  \begin{cases} 2\left(p_u(t) - p_k(t')\right)\tp\mathbf{E}\dot{p}_k(t') & k\neq u,\\ 0 & k=u, \end{cases} \\
\frac{\partial t}{\partial T_{ul}} =&\begin{cases} \frac{j}{\kappa_i} & l=i,\\ 0 & l<i, \end{cases}~~\frac{\partial t'}{\partial T_{ul}} = \begin{cases} \frac{j}{\kappa_i} & l=i,\\ 1 & l<i. \end{cases}
\end{align}
The gradient to $T_{km}$ for any $1\leq m\leq n-1$ can be computed as
\begin{align}
\frac{\partial J_{w}}{\partial T_{km}} &= \sum_{i=1}^{M}\sum_{j=0}^{\kappa_i}\frac{\partial J_{w_k}^{i,j}}{\partial T_{km}},\\
\frac{\partial J_{w_k}^{i,j}}{\partial T_{km}} &= \frac{\partial J_{w_k}^{i,j}}{\partial \mathcal{G}_{w_k}^{i,j}} \frac{\partial \mathcal{G}_{w_k}^{i,j}}{\partial T_{km}},\\
\frac{\partial \mathcal{G}_{w_k}^{i,j}}{\partial T_{km}}&=\frac{\partial \mathcal{G}_{w_k}^{i,j}}{\partial t'}\frac{\partial t'}{\partial T_{km}},\\
\frac{\partial t}{\partial T_{km}} &= -1, ~~~1\leq m \leq n-1.
\end{align}

\paragraph{Uniform Distribution Penalty $J_u$}
Since finite constraint points constrain the collision avoidance, non-uniform distribution increases the possibility of skipping some thin obstacles. Therefore, we penalize the variance of the squared distances between pairs of adjacent constraint points. For more details, please refer to \cite{zhou2021decentralized}.

\section{Evaluation}

In this section, we conduct detailed evaluation tests on the proposed contributions. All programs run on an Intel Core i7-10700KF 5.1GHz CPU.

\subsection{Efficient Multi-Agent Pathfinding}
\label{sec:eva_ecbs}
\begin{figure}[t]
	\centering
	\includegraphics[width=1.0\linewidth]{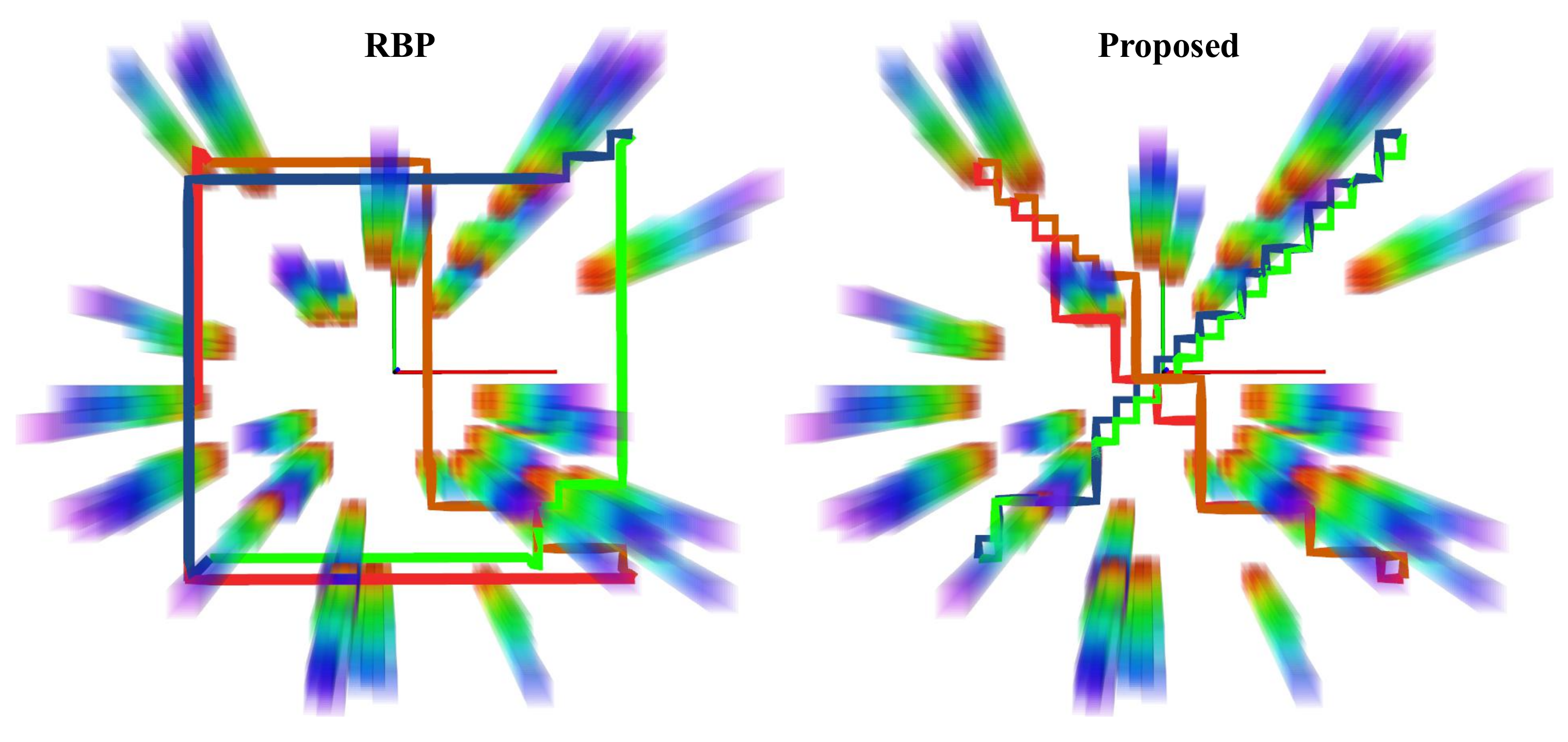}
	\captionsetup{font={small}}
	\caption{The paths searched by the front-end of RBP and the proposed method. The start points of the four agents are the vertices of a $4\times4m$ square. The goal point of each agent is the diagonal position of the start point. The average distance between obstacles is $0.5m$. The resolution of gridmap is $0.1m$. The suboptimal factor $\omega$ is $1.3$. }
	\label{pic:ecbs}
	\vspace{-0.6cm}
\end{figure}

In Fig. \ref{pic:ecbs}, we show that four agents search paths using RBP\cite{park2020efficient} and the proposed method under the same conditions. The initial paths searched by the proposed method are closer to the optimal trajectories, providing a good initial value for the nonlinear optimization problem. The flight time and distance of the proposed method are shorter. The numerical comparison is shown in TABLE \ref{tab:comparisons}.

\begin{figure*}[h]
	\centering
	\includegraphics[width=1.0\linewidth]{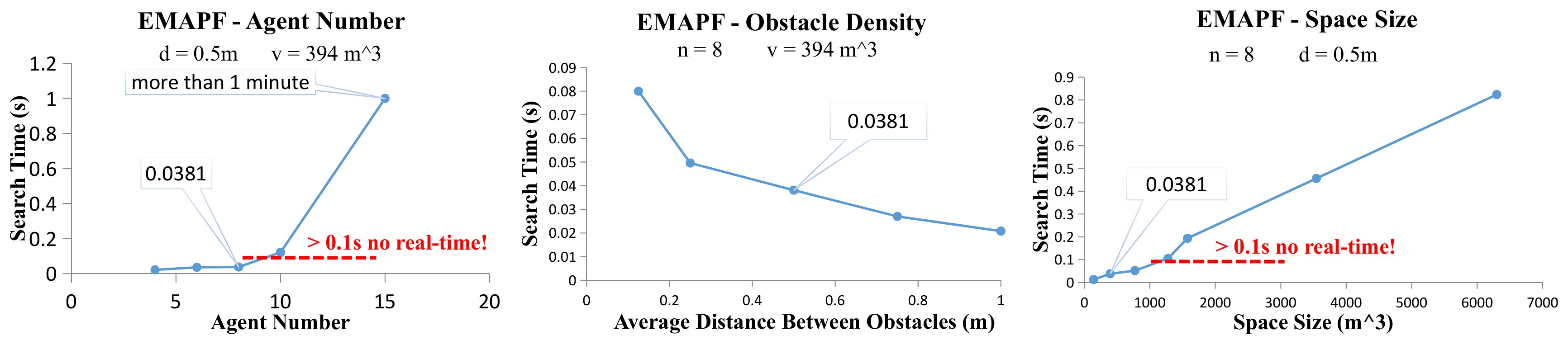}
	\captionsetup{font={small}}
	\caption{The Performance of Efficient Multi-Agent Path Finding(EMAPF); $d$ indicates the average distance between obstacles. $n$ indicates the number of agents. $v$ indicates the space size. The factor $\omega$ is 1.3.}
	\label{pic:EMAPF}
	\vspace{-0.1cm}
\end{figure*}

In Fig. \ref{pic:EMAPF}, to demonstrate the performance of our method, we test the search time from three aspects: the number of agents, the density of obstacles, and the size of the search space. All data shows the average of $10$ runs. We consider the search time of more than $0.1s$ unacceptable for real-time scenarios. 
As a result, we set $n_{max} = 8$ in the Group Planning Criteria. The maximum search space is $1276m^3$.
The average search time is $0.0381s$, which is adequate for real-time usages.

\begin{table*}[]
	\setlength{\abovecaptionskip}{-0.7cm}
	\parbox{.185\textwidth}{\caption{Comparisons in a radius r = 15m circle empty space containing eight agents with radius of $0.25m$. \textbf{int($j^2$)} is time integral of squared jerk, indicating control effort. The units of time and distance are seconds and meters. }
	\label{tab:comparisons}}
    \renewcommand{\arraystretch}{1.45}
	\begin{tabular}{|c|c|c|c|c|c|c|c|c|}
		\hline
		\textbf{\begin{tabular}[c]{@{}c@{}}Offline\\ /Online\end{tabular}} &
		\textbf{Method} &
		\textbf{\begin{tabular}[c]{@{}c@{}}Decentralized\\  Solver Time\end{tabular}} &
		\textbf{\begin{tabular}[c]{@{}c@{}}Centralized\\  Solver Time\end{tabular}} &
		\textbf{\begin{tabular}[c]{@{}c@{}}Replan\\  Times\end{tabular}} &
		\textbf{\begin{tabular}[c]{@{}c@{}}Flight\\  Time\end{tabular}} &
		\textbf{\begin{tabular}[c]{@{}c@{}}Flight\\  Distance\end{tabular}} &
		\textbf{int($j^2$)} &
		\textbf{Safe?} \\ \hline
		\multirow{2}{*}{Offline} &
		\begin{tabular}[c]{@{}c@{}}RBP\\ (bacth size = 1)\end{tabular} &
		\textbackslash{} &
		2.07333 &
		\textbackslash{} &
		42.3539 &
		35.16 &
		0.029 &
		\multirow{6}{*}{Yes} \\ \cline{2-8}
		&
		\begin{tabular}[c]{@{}c@{}}RBP\\ (batch size = 4)\end{tabular} &
		\textbackslash{} &
		1.91592 &
		\textbackslash{} &
		42.3539 &
		35.66 &
		\textbf{0.028} &
		\\ \cline{1-8}
		\multirow{4}{*}{Online} &
		MADER &
		0.007187 &
		\textbackslash{} &
		19573 &
		28.0175 &
		30.28 &
		278.423 &
		\\ \cline{2-8}
		&
		EGO &
		0.000360 &
		\textbackslash{} &
		240 &
		28.4841 &
		31.10 &
		107.079 &
		\\ \cline{2-8}
		&
		DSTO &
		\textbf{0.000154} &
		\textbf{\textbackslash{}} &
		228 &
		21.4581 &
		30.42 &
		44.669 &
		\\ \cline{2-8}
		&
		Proposed &
		0.000353 &
		\textbf{0.02214} &
		\textbf{98} &
		\textbf{20.7576} &
		\textbf{30.06} &
		\textbf{37.407} &
		\\ \hline
	\end{tabular}
	\vspace{-0.5cm}
\end{table*}

\subsection{Comparisons}

In TABLE \ref{tab:comparisons}, we compare the proposed method with RBP\cite{park2020efficient}, Mader\cite{tordesillas2021mader}, EGO (EGO-Swarm)\cite{zhou2020ego}, and DSTO\cite{zhou2021decentralized}. All data is the average of 8 agents in 10 experiment runs. Maximum velocity and acceleration are set to $1.7m/s$ and $6.2m/s^2$. 
We can see from the comparison that RBP has sub-optimal and more conservative trajectories with better smoothness and the longest flight time and distance. Compared with the offline method, the online methods have a replan mechanism and require less solution time. EGO and Mader generate more aggressive trajectories and appear to be more conservative solution spaces because the convex hull constrains the trajectories. EGO, DSTO, and especially Mader need high-frequency replan when the drones cluster.

In Fig. \ref{pic:sim_ben}, we create an extremely challenging environment, consisting of a wall with a narrow gate. There are three drones in each side of the wall with goal points diagonal positions on the other side. We can see that all drones pass through this narrow gate only with the proposed method.

From the comparison, the proposed method shows top-level performance with the shortest flight time, distance, and the fewest replan times. This is attributed to the group planning with the coordination ability.
All simulations and experiments show that the nonlinear optimization problem is always solved in the proposed method, indicating that the safety constraints can be met. However, we still cannot theoretically guarantee that the solution always exists. To ensure safety, we further perform post-check on the optimized trajectories, and only safe trajectories are executed.
If a safety constraint is violated, the planner increases the constraint's weight and solves the problem again. If a safety issue persists, the drone enters an emergency stop mode.

\subsection{The Large-scale Simulation}

\begin{figure}[h]
	\centering
	\includegraphics[width=0.95\linewidth]{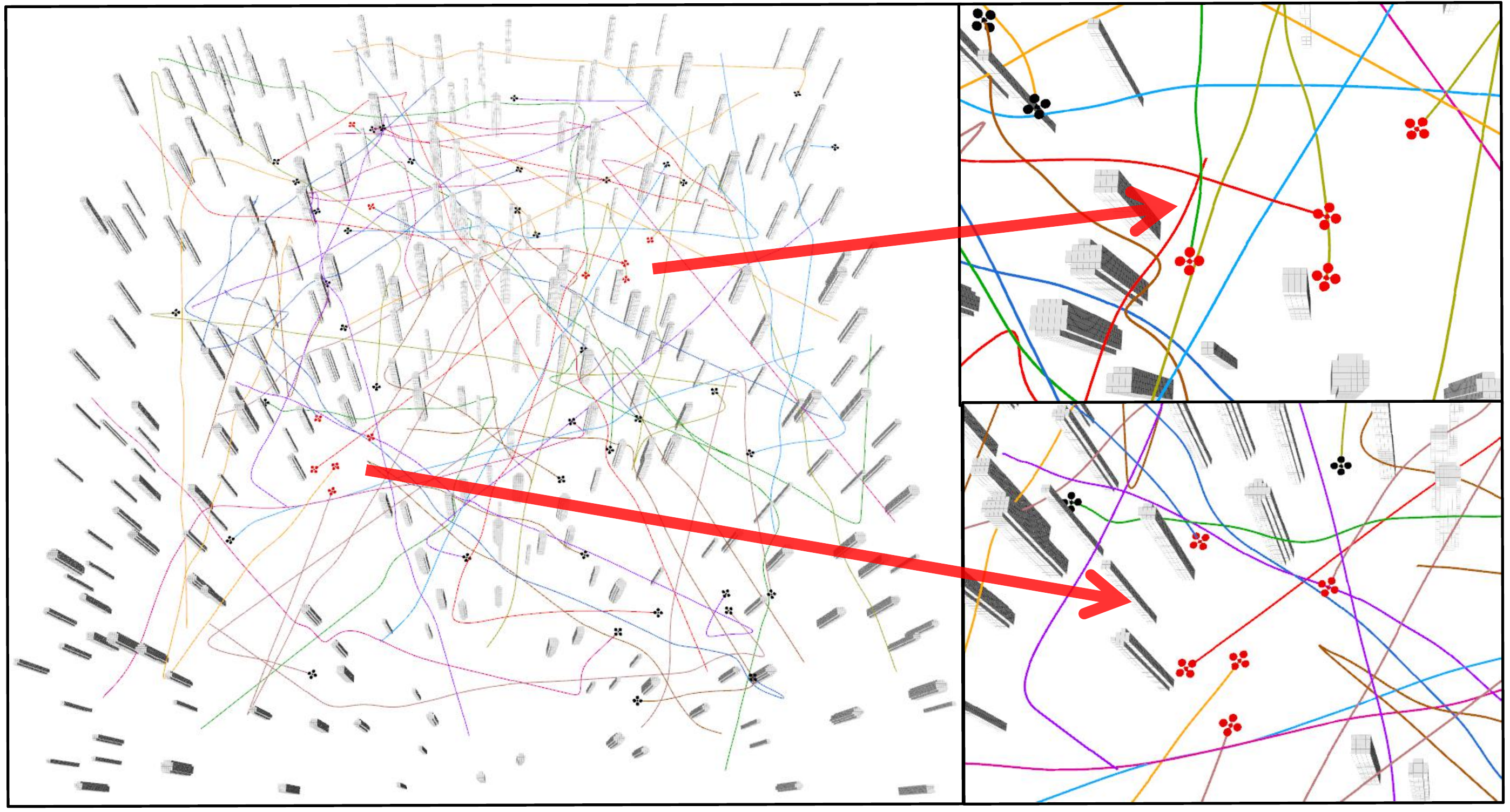}
	\captionsetup{font={small}}
	\caption{Large-scale air traffic simulation; Colored curves indicate the positions that the drones have passed. Maximum velocity and acceleration are set to $1.7m/s$ and $6.2m/s^2$. The average distance between obstacles is $0.5m$.}
	\label{pic:large_scale}
	\vspace{-0.6cm}
\end{figure}

In a $50\times50m$ map, we simulate future air traffic scenarios. The initial positions of 50 drones are generated at random, and each drone must pass through three randomly generated goal positions. Fig. \ref{pic:large_scale} depicts a screenshot of the mission at a certain moment. When multiple drones meet the group planning criteria, group planning is triggered (the drones turn red.). Group planning is triggered $35$ times during the whole mission. The robustness and flexibility of the proposed method in large-scale aerial robot teams are verified. 

\subsection{Discussion On Map Sharing}
\begin{figure}[h]
	\centering
	\includegraphics[width=0.6\linewidth]{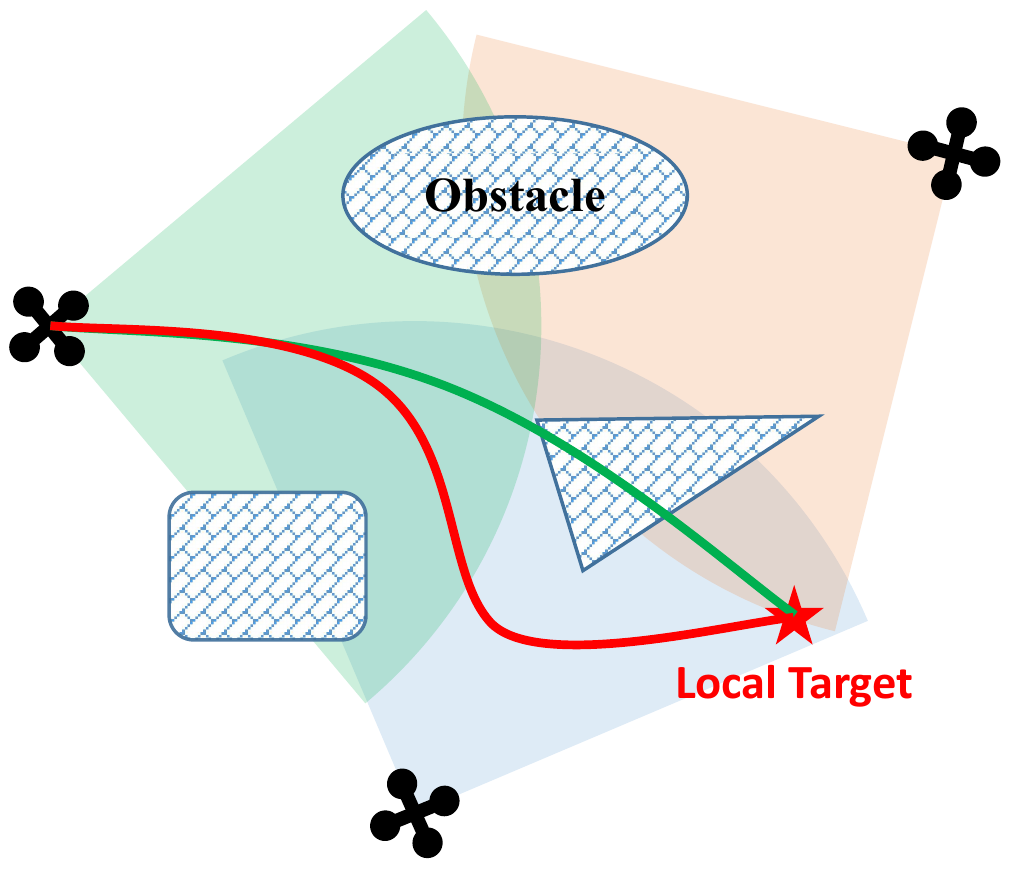}
	\captionsetup{font={small}}
	\caption{The impact of shared maps on trajectory; The red curve indicates the trajectory with shared maps. The green curve indicates the trajectory without shared maps.}
	\label{pic:sharemap}
	\vspace{0.25cm}
\end{figure}
In group planning, each drone's local map is compressed and shared within the group. Therefore, the proposed method can produce safer trajectories, as shown in Fig. \ref{pic:sharemap}.

\section{REAL-WORLD EXPERIMENTS}

\begin{figure*}[ht]

	\centering

	\includegraphics[width=1.0\linewidth]{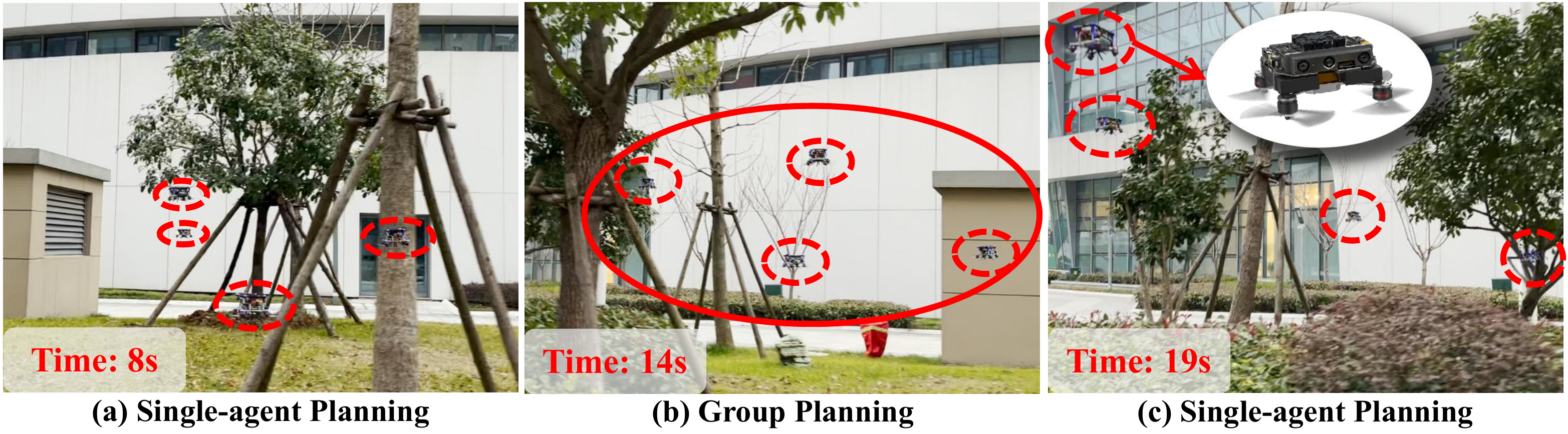}

	\captionsetup{font={small}}

	\caption{Outdoor experiment}

	\label{pic:outdoor}

	\vspace{-0.46cm}

\end{figure*}

\subsection{Indoor}
In the indoor experiments, we place eight drones on two concentric circles with diameters of $10m$ and $4m$ respectively (four drones on each circle). We set the speed limit to $1.0 m/s$. The mission of each drone is to go to the diagonal position of the other circle. In Fig. \ref{pic:indoor}, we set up two environmental experiments: (1) sparse environment (left); (2) dense environment (right); 
In the mission, the entire system completes the mode switch in a manner of ``Single-agent Planning to Group Planning to Single-agent Planning".

\subsection{Outdoor}

In the outdoor experiment, the four drones cross and exchange positions every $8m$ in the longitudinal direction in a dense environment. The entire mission has $5$ cross-flights, and each cross-flight performs the mode switch in a manner of ``Single-agent Planning to Group Planning to Single-agent Planning". In Fig. \ref{pic:outdoor}, the process of one cross-flight is shown. When four drones disperse, each drone performs single-agent planning. When four drones cluster, group planning is triggered. Please watch the video for more information.

\section{CONCLUSIONS AND FUTURE WORK}

In this work, we propose an enhanced decentralized autonomous aerial robot team system with group planning. The system tackles the conflict between planning efficiency and trajectory quality and balances the coordination of individuals and groups. Extensive evaluations demonstrate top-level planning quality and the ability to deploy in a large-scale aerial robot team. Real-world experiments demonstrate the robustness and flexibility of the system. In the future, we will further improve the efficiency and scalability of the multi-agent pathfinding method while considering the kino-dynamic constraints. We will also explore applications in the field of transportation.

\addtolength{\textheight}{-12cm}   


\bibliographystyle{ieeetr}
\bibliography{RAL2022hjl}

\end{document}